\documentclass{article}

\PassOptionsToPackage{numbers, compress, sort}{natbib}
 \usepackage[preprint]{neurips_2026}

\usepackage[utf8]{inputenc} 
\usepackage[T1]{fontenc}    
\usepackage{hyperref}       
\usepackage{url}            
\usepackage{booktabs}       
\usepackage{amsfonts}       
\usepackage{nicefrac}       
\usepackage{microtype}      
\usepackage{xcolor}         
\usepackage{amsmath}
\usepackage{graphicx}
\usepackage{float}
\usepackage{multirow}
\usepackage{tcolorbox}
\tcbuselibrary{skins,breakable}
\providecommand{\up}{\textcolor{teal}{$^{\uparrow}$}}
\providecommand{\dn}{\textcolor{purple}{$^{\downarrow}$}}

\title{Projector Is All You Train}

\author{%
  Nyx Iskandar\thanks{Equal Contribution. Correspondence to: \texttt{nyx@ramenvr.com} or \texttt{nyx@berkeley.edu}.} \\
  Ramen VR \\
  \texttt{nyx@ramenvr.com} \\
  \And
  Saathvik Selvan\footnotemark[1] \space \thanks{Work done while at Ramen VR.} \\
  University of California, Berkeley \\
  \texttt{sselvan@berkeley.edu} \\
  \And
  Slater Victoroff \\
  iph.so \\
  \texttt{s@iph.so} \\
}

\begin{document}

\maketitle

\begin{abstract}
  The typical training process of a multimodal large language model (MLLM) involves adapting both the language model backbone and the projector between the backbone and a modality-specific encoder. We ask whether fine-tuning the backbone of an MLLM is necessary to adapt it to a new modality. Through experiments on 3D MLLMs, we find that training only the projector is sufficient to achieve strong multimodal performance relative to existing baseline models and our jointly trained MLLMs with the same encoder and backbone. We also show that joint training leads to undesirable drift in existing capabilities of the language model, which projector-only training avoids by definition. Furthermore, projector-only training has approximately twice the training sample throughput of joint training. We validate our findings across different language model backbones via 3D classification and captioning benchmarks as well as standard benchmarks evaluating language, vision, and spatial reasoning capabilities.
\end{abstract}

\section{Introduction}

\begin{figure}
  \centering
  \includegraphics[scale=0.15]{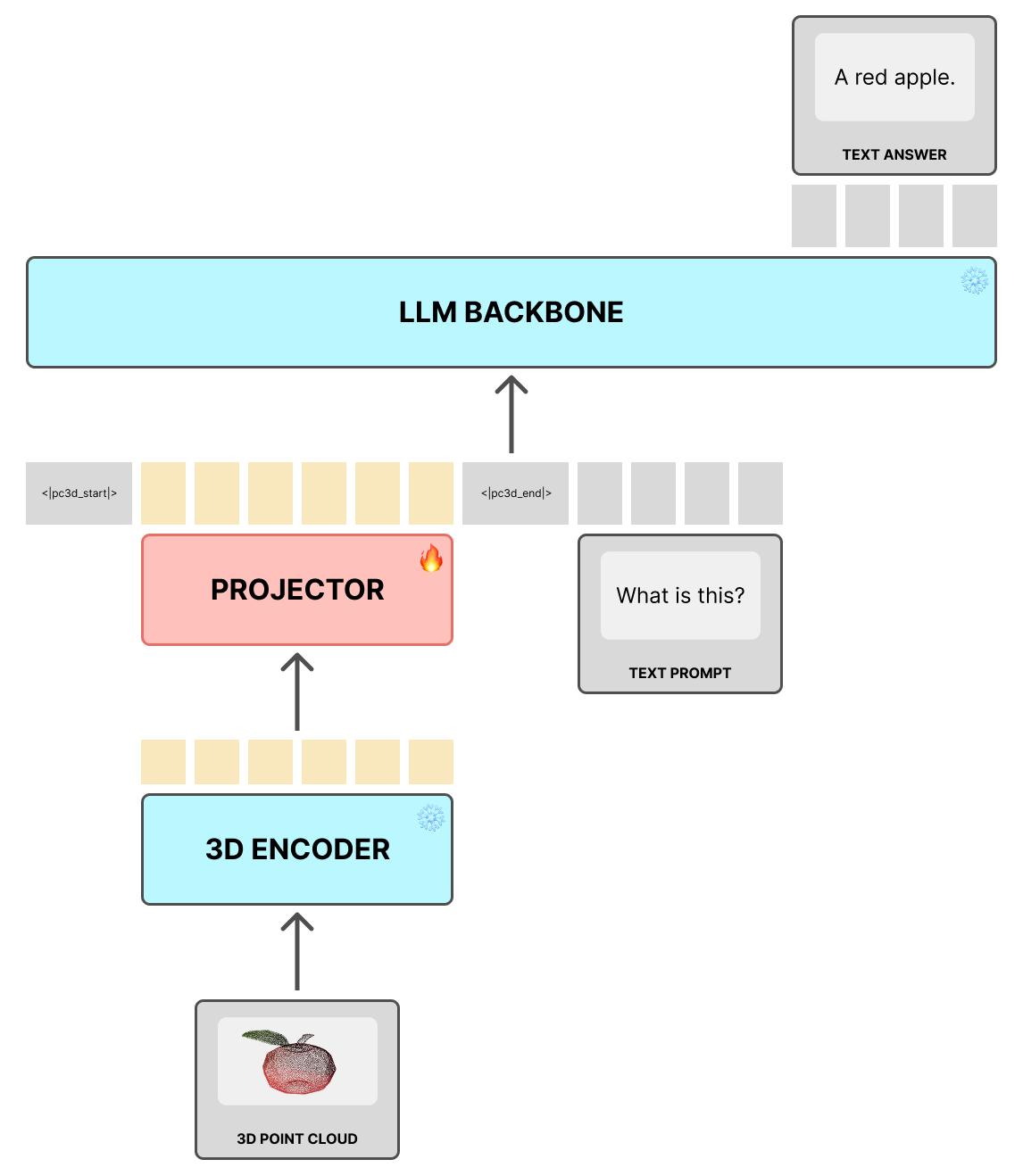}
  \caption{Architecture of 3D MLLMs in this paper. This figure depicts projector-only training, where all parameters are frozen except those of the projector. An alternative training regime is joint training, where all parameters are frozen except those of the projector and LoRA adapters of the LLM backbone.}
  \label{fig:hero}
\end{figure}

Multimodal large language models (MLLMs) extend pretrained language models with representations from additional modalities, such as images, audio, or 3D data \citep{flamingo, blip3o, llava, shapellm, audiogpt}. A common architecture visualized in Figure \ref{fig:hero} maps features from a pretrained modality-specific encoder into the language model's embedding space through a learned projector. Training typically proceeds in two stages: the projector is first trained to align encoder features with the language model (LM), after which both the projector and LM are adapted on multimodal instruction data \citep{pointllm, InstructBLIP, PaLMe, langisnotallyouneed, llava}.

We question whether adapting the LM in this second stage is necessary. For 3D MLLMs, we find that training the projector alone is sufficient to achieve strong multimodal performance. Across multiple LM backbones, models trained with a frozen LM perform comparably to existing baseline models on 3D classification and captioning benchmarks \citep{pointllm, pointllmv2, pointllm-r, minigpt3d}. Furthermore, we independently executed training runs where both the projector and LoRA adapters \citep{lora} of the backbone are jointly trained in a one-stage process using the same dataset as the projector-only training runs. Comparing the 3D classification performance throughout training under matched compute budgets, we find that projector-only training is consistently competitive with joint training when using the same point cloud encoder and LM backbone. These results suggest that training the projector alone can be sufficient to adapt an MLLM to a new modality.

In our experiments, projector-only training is approximately twice as fast as joint training. Moreover, joint training introduces regressions in LM performance across language, vision, and spatial reasoning benchmarks. This drift is guaranteed to be absent in projector-only training by construction, since the backbone weights are never updated.

\section{Related Work}

\textbf{3D MLLMs.} Existing 3D multimodal large language models (MLLMs) differ primarily in their point-cloud representations, modality-alignment strategies, and forms of instruction supervision. ShapeLLM \citep{shapellm} targets interaction-oriented 3D understanding by connecting a LLaMA backbone to ReCon++, a point-cloud encoder extended from ReCon \citep{recon}. PointLLM \citep{pointllm} directly maps features from a Point-BERT encoder pretrained under ULIP-2 \citep{ulip2, pointbert} through an MLP projector into a decoder-only LLM. Its two-stage training first aligns point and language representations using 660k brief object captions and then jointly instruction-tunes the projector and LLM using 70k GPT-4-generated complex interactions; the work also proposes generative 3D classification and captioning benchmarks with human- and GPT-based evaluation. PointLLM-V2 \citep{pointllmv2} expands the dataset to approximately 1.7M point-text samples covering 3D objects contained in Objaverse-XL \citep{objaversexl} in addition to Objaverse \citep{objaverse}, and enriches the instruction space with coordinate-based part-referring questions. PointLLM-V2 also replaces the Vicuna 7B and 13B backbones in PointLLM with Llama-3.1-8B-Instruct. PointLLM-R \citep{pointllm-r} equips PointLLM with explicit chain-of-thought reasoning using PoCoTI, a 55k-sample dataset constructed by first evaluating and refining point-text instructions and then synthesizing geometrically grounded reasoning traces through Human-in-the-Loop Prompt Optimization. Finally, MiniGPT-3D \citep{minigpt3d} emphasizes training efficiency by using a pretrained 2D vision-language model (VLM) as an intermediate semantic bridge between point clouds and an LLM. It combines a four-stage cascaded alignment procedure with a mixture-of-query-experts module, LoRA, and normalization-layer fine-tuning. In this paper, we use the PointLLM family as our baseline for model architecture and dataset, as well as the evaluation strategy it pioneered which has since been adopted in 3D MLLM literature.

\textbf{Point-BERT encoder.} The encoder used by the PointLLM family is Point-BERT, pretrained under ULIP-2 \citep{pointbert, ulip2}. It receives image-text-point contrastive supervision, aligning its representation to a language-adjacent space. Architecturally, Point-BERT adapts the BERT \citep{bert} recipe to point clouds. Input points are partitioned into 512 local patches by farthest-point sampling followed by k-nearest neighbor grouping. After that, a mini PointNet \citep{pointnet} embeds each patch, and the resulting embeddings are passed through a multi-head self-attention transformer. The output is a 513 $\times$ 384 set of tokens (512 patch tokens plus the class token) which we pass to the projector. Point-BERT natively accepts each point as a 6-dimensional vector, where the first three elements represent x-, y-, and z-coordinates, and the last three elements represent the color of that point in RGB format.

\textbf{PointLLM dataset.} We use point clouds released with the PointLLM dataset, produced by surface-sampling each mesh and transferring per-point color from its texture data such that each object is represented as a colored point cloud of $n = 8192$ points with $d=6$ channels, the first three channels being position and the last three RGB \citep{pointllm, objaverse}. For supervision, we use annotations from the PointLLM-V2 dataset, which upgrades the original's text-only supervision with a vision-based pipeline for complex instruction generation \citep{pointllmv2}. This dataset is split into two subsets corresponding to the two-stage training process. The Stage 1 subset contains brief description prompts that ask for a one-line caption of the object supplied by Cap3D \citep{cap3d}; the Stage 2 subset contains more complex prompts spanning long captioning, conversational question-answering, and part-referring. Prompts were either drawn from a fixed collection or generated by GPT-4o \citep{gpt4o}.

\section{Methodology}

This section explains our MLLM architecture and two training regimes. We also outline and motivate our experiment setup.

\subsection{Architecture Design}
\label{sec:architecture}

3D MLLMs are generative models that output a sequence of text tokens given an input point cloud and an input sequence of text tokens. Architecturally, ours consist of a pretrained point cloud encoder $f_{enc}$, a projector $f_{proj}$, and a pretrained LM backbone $f_{lm}$. Only $f_{proj}$ (and, in joint training, low-rank adapters inside $f_{lm}$) is ever optimized, while $f_{enc}$ is frozen in every experiment.

\textbf{3D encoder.} The encoder $f_{enc}$ takes as input a point cloud $P \in \mathbb{R}^{n \times d}$ and outputs a sequence of point features $X \in \mathbb{R}^{m \times c}$, where $n$ is the number of points, $d$ is the feature dimension of each point, $m$ is the number of point features, and $c$ is the dimension of each point feature.

\textbf{Projection module.} The basic purpose of the projector $f_{proj}$ is to map point features $X$ from the output of $f_{enc}$ into the embedding space of $f_{lm}$. This means that $f_{proj}$ takes as input point features $X$ and outputs point tokens $Y \in R^{m \times c'}$, where $c'$ is the dimension of the token embeddings of the corresponding LM. Following the MLP projector configuration recommended by PointLLM-V2, we fix $f_{proj}$ to a 3-layer MLP with hidden dimensions $d_1=1024$ and $d_2=2048$ and GELU \citep{gelu} activations between consecutive linear layers \citep{pointllmv2}.

\textbf{LM backbone.} The language model $f_{lm}$ is a decoder-only transformer \citep{attentionisallyouneed} taking in a sequence of token embeddings $Z \in R^{k \times c'}$, where $k$ is the total number of tokens. $Z$ can contain token embeddings defined in the vocabulary $V$ of the language model as well as point tokens $Y$ outputted by $f_{proj}$. Whenever $Z$ includes $Y$, $Y$ is always surrounded by special tokens \texttt{<|pc3d\_start|>} and \texttt{<|pc3d\_end|>}, which are added to $V$ as input-only tokens. The output of $f_{lm}$ is a sequence of contextual hidden states $\hat{Z} = f_{lm}(Z) \in R^{k \times c'}$. Due to causal masking, each hidden state $\hat{z}_i$ depends only on previous embeddings $Z_{\leq i}$. A linear vocabulary head $f_{vocab}: R^{c'} \rightarrow R^{|V|}$ maps each $\hat{z}_i$ to a logit $l_i = f_{vocab}(\hat{z}_i)$. Under greedy decoding, the predicted next token is $\tilde{z}_{i+1} = \text{arg max}_{w \in V} \text{softmax}(l_i)[w]$. In this paper, the three choices for the LM backbone are Qwen3.5-4B \citep{qwen35}, Qwen3.5-9B \citep{qwen35}, and Llama-3.1-8B-Instruct \citep{llama}.

\subsection{Training Regimes}

We train the relevant parameters of our MLLMs by minimizing the negative log-likelihood, or equivalently the token-level cross-entropy loss, over training token sequences, which is the causal language modeling objective \citep{radford2019language}. For a sequence $x_{1:T}$, the loss minimized to optimize parameters $\theta$ is $\mathcal{L}(\theta) = - \frac{1}{T} \sum{\log{P_{\theta}(x_{t+1} | x_{\leq t})}}$. We ran response-only supervised fine-tuning, masking prompt-token labels such that only assistant-response tokens directly contribute to the loss.

In \textbf{projector-only training}, only the parameters of the projector are optimized. In \textbf{joint training}, both the parameters of the projector and the LoRA adapters \citep{lora} of the LM backbone are optimized, where the adapters are attached to all linear layers of the backbone. Appendix~\ref{app:training} contains more details on the hyperparameters. The dataset used for both training regimes is the same subset of the mixed Stage 1 and Stage 2 data in PointLLM-V2 \citep{pointllmv2} corresponding to objects in Objaverse \citep{objaverse} only. In other words, our training runs do not distinguish between Stage 1 and Stage 2 data when sampling as we do not follow the two-stage training paradigm this dataset was originally constructed for.

\subsection{Experiment Setup}

We execute training runs on MLLMs with different backbones for both projector-only and joint training as shown in Table~\ref{tab:mllm-configs}. Every MLLM is trained under an identical wall-clock time budget of 16 hours on a single A100-80GB GPU. We match by wall-clock time rather than by optimizer steps, since the latter would give joint training approximately $2 \times$ the compute of similar projector-only training runs. This reflects a more realistic GPU budget a practitioner actually faces when deciding which method to use. The 16 hours are measured from the first optimizer step, excluding dataset preprocessing and model loading time. All runs use identical batch and sequence settings. We report additional training details, including hyperparameters and throughput, in Appendix~\ref{app:training}.

\begin{table}[h]
  \caption{Configurations of MLLM variants. All variants are trained for the full 16 GPU hours. We save checkpoints at 2, 4, 8, 12, and 16 hours.}
  \label{tab:mllm-configs}
  \centering
  \begin{tabular}{llcc}
    \toprule
    ID        & LM Backbone & Projector Trained & LoRA Fine-tuned \\
    \midrule
    P-Llama8B & Llama-3.1-8B-Instruct & Yes & No      \\
    J-Llama8B & Llama-3.1-8B-Instruct & Yes & Yes  \\
    P-Qwen4B  & Qwen3.5-4B & Yes & No     \\
    J-Qwen4B  & Qwen3.5-4B & Yes & Yes      \\
    P-Qwen9B  & Qwen3.5-9B & Yes & No  \\
    J-Qwen9B  & Qwen3.5-9B & Yes & Yes     \\
    \bottomrule
  \end{tabular}
\end{table}

Our experiments vary the language model backbone but not the encoder. This is to control our experiments and make the trained MLLMs comparable to the baseline PointLLM family which all use the Point-BERT encoder. Furthermore, the encoder remains perpetually frozen under both training regimes; on the other hand, the backbones are ablated as they are updated in joint training. By and large, we are more interested in validating that our findings hold for different language models.

\textbf{3D understanding evaluations.} We evaluate model checkpoints at 2-, 4-, 8-, 12-, and 16-hour thresholds against benchmarks assessing 3D understanding \citep{pointllm-r, pointllm}. These include generative zero-shot 3D object classification benchmarks and a 3D object captioning benchmark first proposed by PointLLM. The classification benchmarks use point clouds from ModelNet40 \citep{modelnet40}, Objaverse \citep{objaverse} and, since PointLLM-R, OmniObject3D \citep{omniobject3d}. The captioning benchmark uses Objaverse point clouds. For evaluations, the classification benchmarks use LLM-as-a-Judge to score output generation correctness; in this paper, the judge used is GPT-5.6 Luna \citep{gpt56} and existing models are re-run and re-scored under said judge. The Objaverse captioning benchmark also uses LLM-as-a-Judge that is given multi-view images of the 3D object to determine a correctness and hallucination score for the generated open-ended caption, aggregated to a precision score as computed by PointLLM; in this paper, we use an ensemble of judges, namely GPT-5.6 Luna \citep{gpt56}, Claude Haiku 4.5 \citep{haiku45}, and Gemini 3.5 Flash-Lite \citep{gemini35}. More details on the evaluation configuration, including the judge prompts, are provided in Appendix~\ref{app:eval-procedure}.

\textbf{LM backbone evaluations.} To determine the extent to which existing capabilities of the LM drift after LoRA fine-tuning for joint training variants, we load and merge the trained LoRA adapters into the base LM weights and run standard benchmarks against the LM backbone. These include language benchmarks like MMLU-Pro \citep{mmlupro} and WinoGrande \citep{winogrande}, vision benchmarks like MMMU-Pro \citep{mmmupro} and BabyVision \citep{babyvision}, and spatial intelligence benchmarks like ERQA \citep{erqa} and LingoQA \citep{lingoqa}. We also ran these benchmarks against the base LM itself to obtain a verified baseline; the base LM corresponds to the backbone of projector-only-trained MLLMs. Since Llama-3.1-8B-Instruct is a text-only LLM, we only ran language capability benchmarks against it.

All in all, we aim to show three things through our experiments. First, by evaluating our trained MLLMs on classification and captioning benchmarks, we quantitatively show that projector-only-trained MLLMs obtain 3D understanding scores comparable to existing methods. Second, by comparing classification accuracy scores over time, we show that there is generally no evidence that joint training outperforms projector-only training at any GPU-hour threshold. We validate this up to 16 hours, at which point we stop training. Finally, we show through evaluating jointly trained backbones and base LLMs against standard benchmarks that joint training leads to drift in language, vision, and spatial reasoning capabilities to varying degrees.

\section{Evaluations and Results} \label{sec:evaluations-and-results}

We report and analyze our 3D understanding and LM backbone benchmark results in this section.

\subsection{3D Understanding}

We report our 3D classification and captioning benchmark results for all Table~\ref{tab:mllm-configs} variants to compare projector-only training against existing models and against joint training.

\subsubsection{Competitive With Existing Methods}

\begin{figure}[h]
  \centering
  \includegraphics[scale=0.4]{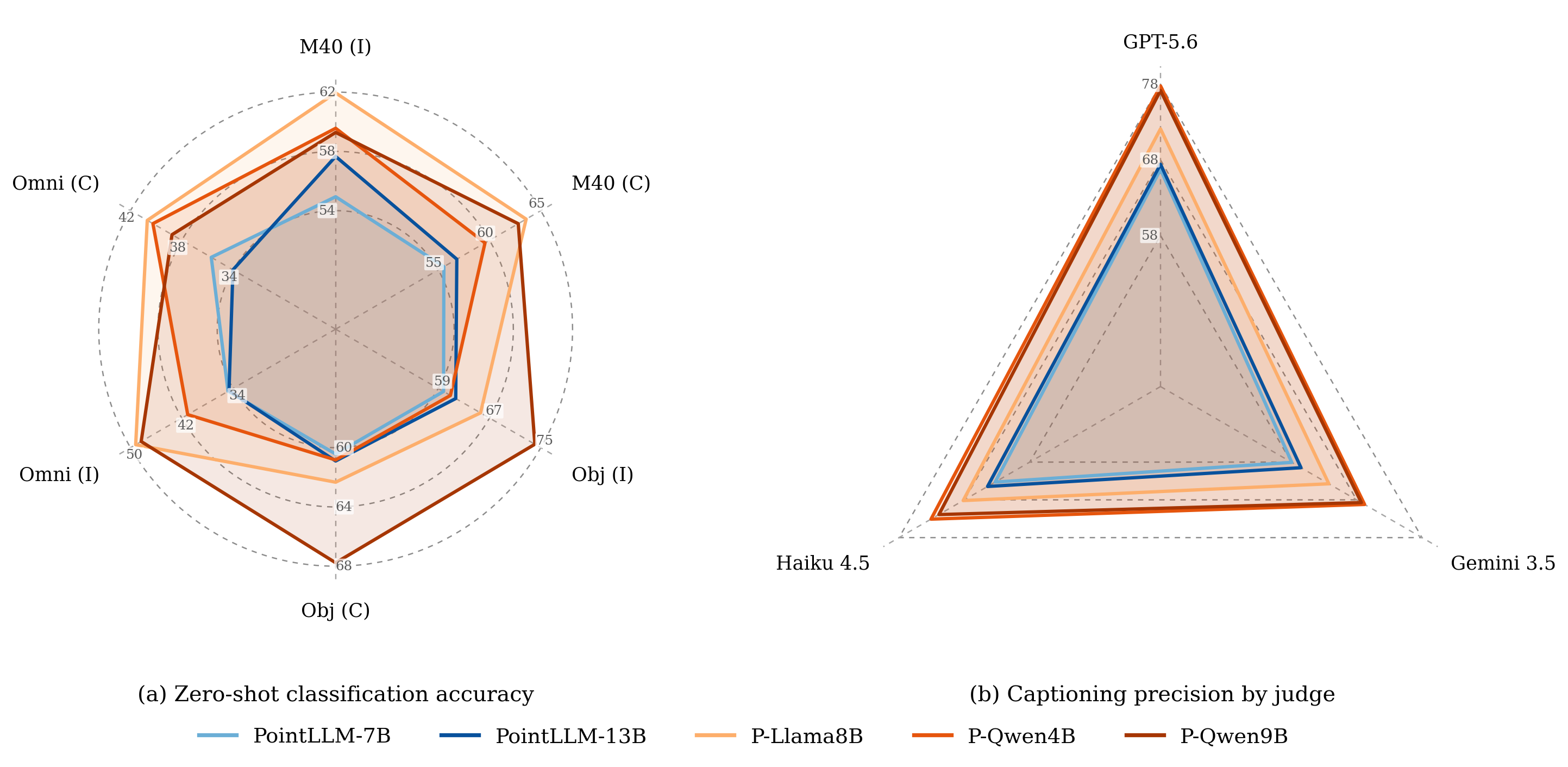}
  \caption{Evaluation results of projector-only-trained MLLMs against baseline PointLLM models. Chart (a) shows generative 3D object classification results on ModelNet40 (M40.), Objaverse (Obj.), and OmniObject3D (Omni.) objects under a zero-shot setting. There are two prompt types per benchmark: an instruction-style prompt (I, ``What is this?'') and a completion-style prompt (C, ``This is an object of''). Each entry reports accuracy judged by GPT-5.6 Luna \citep{gpt56}. Chart (b) shows the aggregate precision score on 3D object captioning tasks under the three judge LLMs. More details are found in Appendix~\ref{app:eval-procedure}.}
  \label{fig:against-existing}
\end{figure}

Figure~\ref{fig:against-existing} compares projector-only-trained MLLMs against the baseline PointLLM models on 3D classification and captioning tasks. We independently run these evaluations on existing models as our judge LLM differs from past papers. We do not report PointLLM-V2 scores due to the lack of publicly available weights. Tables~\ref{tab:results-3dund} and \ref{tab:caption-16h} show the exact scores for classification and captioning respectively, and additional tables in Appendix~\ref{app:add-results} include the scores of more existing models.

All projector-only-trained MLLMs achieve scores comparable to or higher than those of PointLLM \citep{pointllm}. In fact, all PointLLM scores lie below the 50$^{\text{th}}$ percentile of each evaluation vertical. In Appendix~\ref{app:add-results}, projector-only scores are also competitive when compared against improved methods, namely PointLLM-R and MiniGPT-3D \citep{pointllm-r, minigpt3d}. The most fair comparison against the PointLLM family would be one made with P-Llama8B, given that the architecture is the exact replica of that of PointLLM-V2 \citep{pointllmv2}; our checkpoint outperforms both PointLLM models conclusively.

Comparing between projector-only backbones, we observe different trends across both task types. While classification scores tend to improve with increasing backbone parameter count, higher captioning scores are obtained by MLLMs with native VLM backbones (Qwen family) rather than the text-only LLM backbone (Llama). We hypothesize that this is because the VLM backbones have been trained on more data that resemble captions, allowing them to output generations more favorably scored by the judge LLMs for captioning tasks.

\begin{table}[h]
    \caption{Generative 3D object classification results on ModelNet40 (M40.), Objaverse (Obj.), and OmniObject3D (Omni.) under a zero-shot setting as reported in Figure~\ref{fig:against-existing} (a).}
    \label{tab:results-3dund}
    \centering
    \begin{tabular}{lcccccc}
    \toprule
    Model & M40. (I) & M40. (C) & Obj. (I) & Obj. (C) & Omni. (I) & Omni. (C) \\
    \midrule
    PointLLM-7B \cite{pointllm}  & 54.94 & 55.55 & 59.83 & 60.43 & 34.78 & 35.70 \\
    PointLLM-13B \cite{pointllm} & 57.66 & 56.81 & 61.73 & 60.90 & 34.63 & 34.01 \\
    \textbf{P-Llama8B}  & 61.95 & 63.57 & 65.60 & 62.33 & 49.19 & 40.69 \\
    \textbf{P-Qwen4B}  & 59.56 & 59.56 & 60.90 & 60.83 & 41.07 & 40.25 \\
    \textbf{P-Qwen9B}  & 59.28 & 62.80 & 74.07 & 67.77 & 48.33 & 38.76 \\
    \midrule
    Mean & 58.68 & 59.66 & 64.43 & 62.45 & 41.60 & 37.88 \\
    \bottomrule
    \end{tabular}
\end{table}

\begin{table}[h]
    \caption{3D object captioning results on Objaverse as reported in Figure~\ref{fig:against-existing} (b). C refers to correctness, H to hallucination, and P to precision (aggregate of C and H).}
    \label{tab:caption-16h}
    \centering
    \begin{tabular}{lcccccccccc}
      \toprule
       & \multicolumn{3}{c}{GPT-5.6 Luna}
       & \multicolumn{3}{c}{Claude Haiku 4.5}
       & \multicolumn{3}{c}{Gemini 3.5 Flash-Lite}
       & \\
      \cmidrule(lr){2-4} \cmidrule(lr){5-7} \cmidrule(lr){8-10}
      Model & C & H $\downarrow$ & P
            & C & H $\downarrow$ & P
            & C & H $\downarrow$ & P \\
      \midrule
      PointLLM-7B \citep{pointllm}  & 4.26 & 2.12 & 66.75 & 3.37 & 1.95 & 63.30 & 2.75 & 1.99 & 58.08 \\
      PointLLM-13B \citep{pointllm} & 4.35 & 2.10 & 67.43 & 3.47 & 1.91 & 64.46 & 2.82 & 1.92 & 59.49 \\
      \textbf{P-Llama8B}    & 5.61 & 2.17 & 72.12 & 3.76 & 1.75 & 68.19 & 3.29 & 1.87 & 63.75 \\
      \textbf{P-Qwen4B}     & 6.41 & 1.83 & 77.81 & 4.28 & 1.57 & 73.11 & 3.72 & 1.65 & 69.22 \\
      \textbf{P-Qwen9B}     & 6.58 & 1.95 & 77.19 & 4.30 & 1.68 & 71.90 & 3.76 & 1.71 & 68.72 \\
      \midrule
      Mean & -- & -- & 72.26 & -- & -- & 68.19 & -- & -- & 63.85 \\
      \bottomrule
    \end{tabular}
\end{table}

\textbf{Curriculum learning ablation.} In our training runs, all data is mixed together and sampled without discriminating between Stage 1 and Stage 2. As an ablation, we trained an MLLM first on Stage 1 data and then on Stage 2 data, with an otherwise equivalent configuration to P-Qwen4B. This ablation variant is trained for the full 16 hours, with the proportion of Stage 1 data to Stage 2 data equal to that of all other variants. The scores for this ablation are reported in extended Tables \ref{tab:results-3dund-extension} and \ref{tab:caption-16h-extension}. Comparing these results to those of P-Qwen4B, mixing data leads to better performance in some benchmarks but worse in others, thus no definitive evidence exists to argue for nor against it.

\textbf{Removing trained LoRA adapters from jointly trained MLLMs.} We also ran evaluations against jointly trained MLLMs without the trained LoRA adapters loaded. Using only the trained projectors with the base LLM, the scores generally degrade by varying degrees, as reported in extended Tables \ref{tab:results-3dund-extension} and \ref{tab:caption-16h-extension}. These degraded scores are lower than the corresponding projector-only-trained scores, demonstrating that the drift in existing language model capability reported in Section~\ref{sec:existing-caps} is a necessary tradeoff for jointly trained MLLMs to be competitive with projector-only-trained MLLMs in 3D understanding capability. Interestingly, some scores still exceed PointLLM's, which suggests that the projector carries much of the 3D understanding capability even during joint training.

Through these results, we quantitatively demonstrate that projector-only training leads to competitive 3D understanding capabilities. This is also an important finding from the perspective of comparing projector-only training against joint training (i.e., comparing P-* against J-* variants): we now establish that any findings regarding this comparison cannot be attributed to under-training.

\subsubsection{Competitive With Joint Training}

\begin{figure}[h]
  \centering
  \includegraphics[scale=0.25]{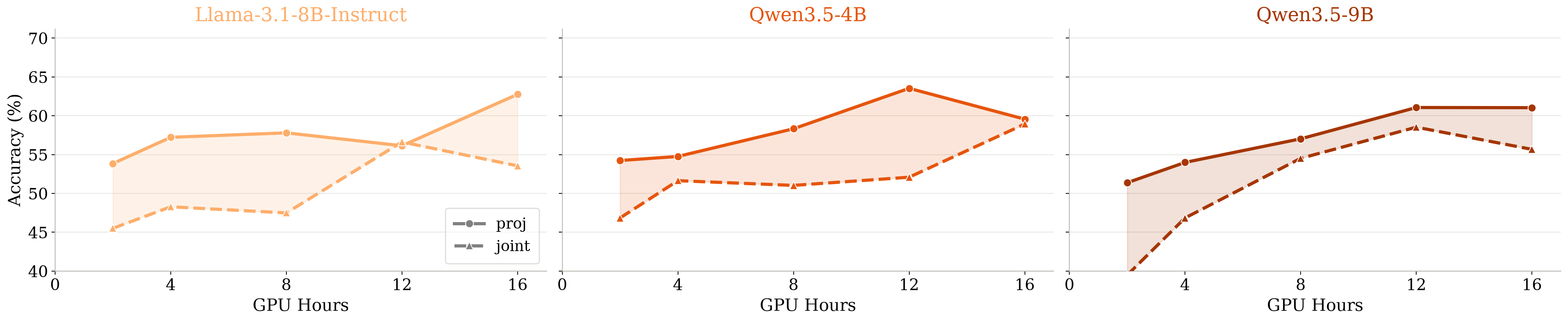}
  \includegraphics[scale=0.25]{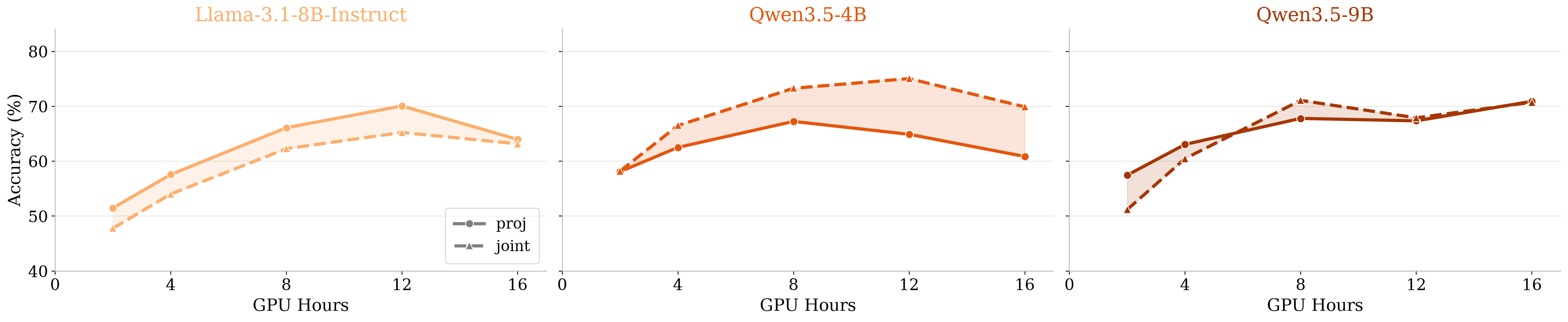}
  \caption{M40. (top) and Obj. (bottom) mean accuracy against GPU hours. Projector-only training generally scores higher than joint training across all GPU hours.}
  \label{fig:acc-over-time}
\end{figure}

\begin{figure}[h]
  \centering
  \includegraphics[scale=0.25]{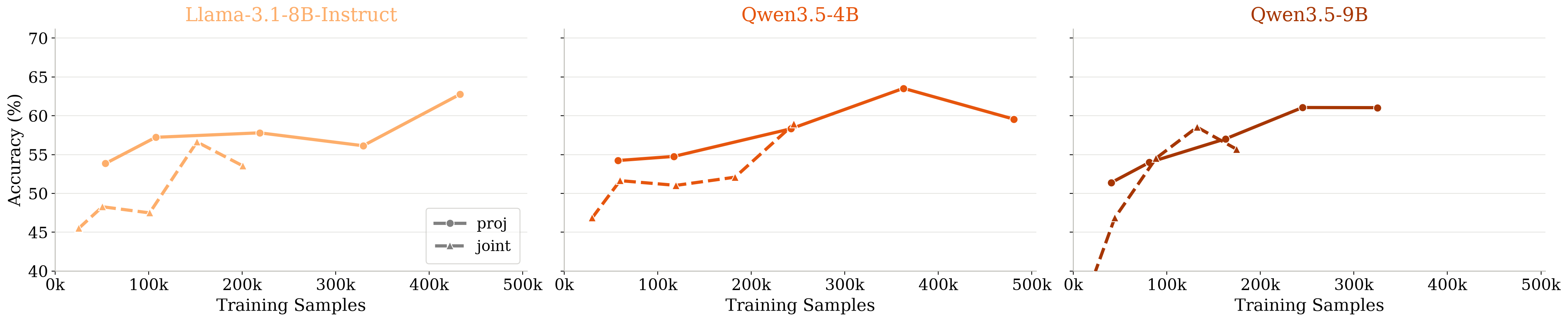}
  \includegraphics[scale=0.25]{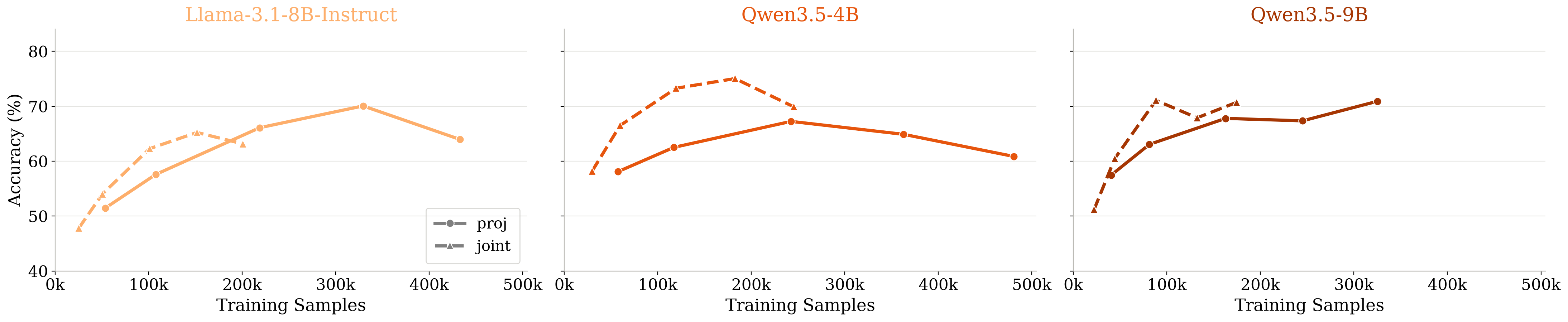}
  \caption{M40. (top) and Obj. (bottom) mean accuracy against training samples. Plotted points in these charts correspond to the same GPU-hour thresholds that in turn correspond to the plotted points in Figure \ref{fig:acc-over-time}. Projector-only training sees more samples than joint training at each GPU-hour threshold as it is approximately twice as fast.}
  \label{fig:acc-over-samples}
\end{figure}

Figure~\ref{fig:acc-over-time} plots accuracy scores of two 3D classification benchmarks against training GPU hours, specifically the averages of the I and C variants of the ModelNet40 and Objaverse benchmarks from Table~\ref{tab:results-3dund}. We collect evaluation data for all variants at various GPU-hour thresholds, namely 2, 4, 8, 12, and 16 hours. As a companion, Figure~\ref{fig:acc-over-samples} plots accuracy scores against number of training samples seen, where each plotted point corresponds to the respective GPU-hour threshold. These figures compare projector-only and joint training performance across different language model backbones, with exact scores reported in Table \ref{tab:acc-over-time-numbers}.

Generally, projector-only training scores higher at most thresholds, with the difference in accuracy scores being negligible at several time thresholds. Especially for M40., projector-only consistently outperforms joint. These results establish that, under similar conditions, projector-only training reaches competitive 3D understanding across all three backbones. Joint training is therefore not necessary for 3D understanding to emerge.

Since projector-only training runs at approximately double the speed of joint training, it sees more samples within the same wall-clock time. Training speed also depends on the forward pass cost through the backbone, which accounts for the differences in sample counts across backbones but cancels within each P/J pair. Reading Figure~\ref{fig:acc-over-samples} at matched sample counts, projector-only training is more sample efficient when evaluated under M40., but less sample efficient under Obj. However, such as for P-Llama8B on Obj., projector-only accuracy is still rising even as its joint counterpart stops improving, so part of joint training's apparent sample advantage reflects earlier saturation rather than better learning. In any case, sample efficiency is secondary in practice since a practitioner with a fixed GPU budget is constrained by compute rather than by samples, given that neither regime exhausted the training pool in our runs.

We propose a hypothesis regarding the reason for the outlier trend in P-Qwen4B and J-Qwen4B on Obj. In Figures \ref{fig:acc-over-time} and \ref{fig:acc-over-samples}, it is quite visible that both P-Qwen4B and J-Qwen4B have declined in performance prior to the 16-hour mark, suggesting that there could be a better hyperparameter combination (e.g., different seed for the dataloader) that could have prevented early saturation and led to a narrower difference in performance between projector-only and joint. Having said that, it is interesting to see that J-Qwen4B outperforms both J-Llama8B and J-Qwen9B while P-Qwen4B lags behind both P-Llama8B and P-Qwen9B.

\subsection{Existing Language, Vision, and Spatial Reasoning Capabilities} \label{sec:existing-caps}

Catastrophic forgetting in MLLMs is a known phenomenon \citep{zhai2023investigatingcatastrophicforgettingmultimodal}. In this paper, none of our backbones have native 3D capabilities in their base versions as they are not equipped with 3D encoders; Qwen3.5-4B and Qwen3.5-9B are VLMs, while Llama-3.1-8B-Instruct is text-only. This means that catastrophic forgetting cannot occur for the 3D modality, rather it is only relevant for existing capabilities in language, vision, and image-based spatial reasoning. Since the projector maps solely features from the 3D point cloud encoder, and the LM backbone in projector-only training is unmodified, catastrophic forgetting in this work only applies to joint training.

To determine the extent of drift in existing capabilities, we evaluate the backbones of our jointly trained MLLMs at the 16-hour checkpoint against standard language, vision, and spatial reasoning benchmarks. We also evaluate the backbones of projector-only-trained MLLMs, which are equivalent to their base LLMs that have not been fine-tuned on our dataset. We report these results in Table~\ref{tab:results-std}.

\begin{table}[h]
    \caption{Results for language, vision, and spatial reasoning benchmarks. The LoRA fine-tuned backbones are compared against the corresponding base backbone.}
    \label{tab:results-std}
    \centering
    \setlength{\tabcolsep}{4pt}
    \begin{tabular}{lcccccc}
    \toprule
    & \multicolumn{2}{c}{Llama-3.1-8B-Instruct} & \multicolumn{2}{c}{Qwen3.5-4B} & \multicolumn{2}{c}{Qwen3.5-9B} \\
    \cmidrule(lr){2-3} \cmidrule(lr){4-5} \cmidrule(lr){6-7}
    Benchmark & P & J & P & J & P & J \\
    \midrule
    \multicolumn{7}{l}{\textbf{Language}} \\
    MMLU-Pro \citep{mmlupro}           & 37.32 & 11.21\dn & 45.04 & 45.48 & 51.38 & 51.30 \\
    MMLU-Redux \citep{mmluredux}         & 60.37 & 22.83\dn & 69.13 & 70.53 & 74.57 & 74.27 \\
    GPQA Diamond  \citep{gpqa}      & 33.33 & 24.24\dn & 33.33 & 41.41\up & 45.96 & 49.49\up \\
    IFEval \citep{ifeval}             & 72.46 & 10.35\dn & 80.41 & 38.82\dn & 83.55 & 67.84\dn \\
    IFBench \citep{ifbench}           & 26.33 & 16.33\dn & 29.33 & 17.67\dn & 32.67 & 28.33\dn \\
    GSM8K \citep{gsm8k}              & 86.96 &  0.61\dn & 90.90 & 81.96\dn & 92.34 & 92.34 \\
    WinoGrande \citep{winogrande}         & 61.40 & 49.57\dn & 65.59 & 68.35\up & 74.19 & 73.24 \\
    OpenBookQA \citep{openbookqa}         & 81.60 & 27.60\dn & 86.20 & 86.90 & 90.30 & 92.20\up \\
    HumanEval \citep{humaneval} & 64.02 &  0.00\dn & 82.93 & 71.95\dn & 84.76 & 80.49\dn \\
    \midrule
    \multicolumn{7}{l}{\textbf{Vision}} \\
    MMMU \citep{mmmu}               & -- & -- & 49.44 & 54.45\up & 54.23 & 59.02\up \\
    MMMU-Pro \citep{mmmupro}           & -- & -- & 32.60 & 37.23\up & 42.72 & 43.41 \\
    MMMU-Pro Vision \citep{mmmupro}   & -- & -- & 31.45 & 34.16\up & 40.46 & 40.64 \\
    MMStar \citep{mmstar}             & -- & -- & 53.27 & 61.67\up & 65.87 & 65.67 \\
    BabyVision \citep{babyvision}         & -- & -- & 19.07 & 14.43\dn & 16.49 & 15.98 \\
    RealWorldQA \citep{realworldqa}        & -- & -- & 74.12 & 69.54\dn & 75.42 & 76.08 \\
    \midrule
    \multicolumn{7}{l}{\textbf{Spatial}} \\
    ERQA \citep{erqa}               & -- & -- & 45.25 & 42.25\dn & 45.75 & 44.00\dn \\
    
    EmbSpatialBench \citep{embspatialbench}    & -- & -- & 75.14 & 75.16 & 76.59 & 76.95 \\
    RefSpatialBench \citep{refspatialbench}    & -- & -- & 20.94 &  1.81\dn & 38.27 & 27.80\dn \\
    LingoQA \citep{lingoqa}            & -- & -- & 70.40 & 58.80\dn & 75.00 & 68.40\dn \\
    \bottomrule
    \end{tabular}
\end{table}

For language capability, there is an overall degradation in performance, especially for the Llama backbone. Analyzing the model outputs, we find the significant drift for the Llama backbone is due to its inability to form coherent text for open-ended questions, which is the format of IFEval \citep{ifeval}, IFBench \citep{ifbench}, and GSM8K \citep{gsm8k}. The Llama backbone also fails to follow instructions to generate code, hence collapsing to a zero score on HumanEval \citep{humaneval}.

There is interestingly a general improvement in vision scores, relevant only for the Qwen backbones, though this quantitative result is misleading. The benchmarks for which the scores noticeably improved, namely MMMU \citep{mmmu}, MMMU-Pro \citep{mmmupro}, MMMU-Pro Vision \citep{mmmupro}, and MMStar \citep{mmstar}, all contain multiple-choice questions that are scored by renormalizing the model's likelihoods over only the answer options. This means the evaluations do not rely on decoded text generations, and that the resulting score is invariant to the absolute probability the model places on the correct answer. Investigating the outputs to open-ended benchmarks like BabyVision \citep{babyvision} and RealWorldQA \citep{realworldqa}, we see instruction-following failure in jointly trained backbones (e.g., responding to a question asking for a number with a caption-like answer) that may have arisen due to the narrow distribution of text types in the fine-tuning 3D dataset. These findings indicate that vision capability has not improved, and that the improved scores are actually misleading due to the mechanism of the evaluation.

The same conclusion can be drawn from the spatial results. Note that these benchmarks do not natively feed 3D inputs into the models; rather, they use 2D images paired with questions that are subcategorized as spatial reasoning questions. Jointly trained backbones again often fail to follow instructions appropriately, such as responding with a description of the object rather than outputting the coordinates of the object of interest. We again attribute this collapse to the dataset used to fine-tune the MLLMs for 3D understanding.

Overall, the degradation in the existing capabilities of the LM backbones is significant enough to dissuade practitioners from conducting joint training to build general MLLMs. Since projector-only training evidently achieves competitive 3D understanding capabilities, the effort to optimize hyperparameters for joint training to prevent backbone drifts may not be warranted.

\section{Conclusion}

In this work, we show that adapting the language model in training multimodal large language models is unnecessary. By conducting training runs and evaluations across different language model backbones, we show that projector-only training results in 3D MLLMs with capabilities comparable to those of existing baseline 3D MLLMs, and that projector-only training is also competitive with joint training under matched compute budgets. Hence, training the projector alone is sufficient for building an MLLM.

Projector-only training provides two demonstrable advantages over joint training. Firstly, projector-only training has a higher iteration speed than joint training, allowing it to see more training samples during the same wall time budget. Secondly, as the LM backbone in projector-only training is frozen, it experiences absolutely no drift in its existing capabilities. Through relevant benchmarks, we show that joint training, on the other hand, does cause a noticeable degradation in existing capabilities of the LM backbone due to supervised fine-tuning.

As an attempt to understand why projector-only training is so effective, we draw parallels to prompt engineering. We hypothesize that optimizing the output of the projector is akin to searching for the best sequence of discrete tokens that yields the best performance in a large language model for a particular task. Despite freezing the backbone, the MLLM learns to answer 3D understanding questions correctly, which means that the projector uses its input features from the encoder to elicit correct token distributions from the backbone. We leave interpretability to future work.

We are excited about the possibility of a more scalable, modular approach to training general MLLMs for all modalities with projector-only training. Since the LM backbone need not be fine-tuned, the same LM can serve as the backbone for multiple learned projectors and their corresponding encoders for multiple modalities, each being able to be independently trained. While our work is limited to 3D point cloud encoders, there has been prior work contrasting catastrophic forgetting between linear (projector-only) and LoRA fine-tuning that implicitly shows that projector-only fine-tuning still leads to 2D image understanding capability \cite{zhai2023investigatingcatastrophicforgettingmultimodal}. We believe that testing projector-only training on MLLMs of different modalities is a valuable future research direction towards modular MLLM training.

\begin{ack}
Nyx Iskandar and Saathvik Selvan completed this work as Research Engineer and Research Fellow at Ramen VR, respectively. Slater Victoroff is the advisor of the project. The authors would like to thank Andy Tsen and the Ramen VR team for their support and funding.
\end{ack}

\bibliography{references}


\newpage

\appendix

\section{Training Configuration}
\label{app:training}

\subsection{Input Representation and Model}

Each object enters the encoder as $n = 8192$ points with $d = 6$ channels. Positions are normalized to the unit sphere and RGB values are scaled to $[-1, 1]$.

The projector is a 3-layer MLP mapping $c \rightarrow 1024 \rightarrow 2048 \rightarrow c'$ with GELU applied between consecutive linear layers and no activations on the output. The projector is trained and stored in \texttt{float32} while the backbone runs in \texttt{bfloat16}, since the projector is small enough that full precision costs very little. The two boundary tokens \texttt{<|pc3d\_start|>} and \texttt{<|pc3d\_end|>} are appended to the vocabulary and initialized to the mean of the base embedding matrix and frozen. Since our models are not required to output these tokens, we are able to freeze the entire backbone during projector-only training, allowing us to guarantee zero drift of the base LM capabilities.

\begin{table}[h]
\small
\centering
\caption{Projector and LoRA-adapter parameter counts, and projector input/output shapes, for each backbone.}
\label{tab:projparams}
\begin{tabular}{lccccc}
\toprule
Backbone & Encoder dim $c$ & Tokens $m$ & Backbone dim $c'$ & Proj. params & LoRA params \\
\midrule
Llama-3.1-8B-Instruct  & 384  & 513  & 4096  & $10.89$M & $41.94$M \\
Qwen3.5-4B   & 384  & 513  & 2560  & $7.74$M  & $32.46$M \\
Qwen3.5-9B   & 384  & 513  & 4096  & $10.89$M & $43.28$M \\
\bottomrule
\end{tabular}
\end{table}

\subsection{Data and Sampling}

\paragraph{Corpus construction.} Training rows are drawn from the union of the Stage 1 and Stage 2 subsets of the PointLLM-V2 dataset~\citep{pointllmv2}, restricted to objects appearing in Objaverse~\citep{objaverse} and excluding Objaverse-XL \citep{objaversexl}. After filtering, the pool contains 1,368,302 rows over 661,375 unique objects, containing 661,375 Stage 1 rows and 706,927 Stage 2 rows. Stage 1 contains exactly one brief description row per object, while Stage 2 contains several rows per object, divided into 77,440 captioning, 310,415 question-answering, and 319,072 referring rows.

\paragraph{Sampling.} Rows are sampled uniformly without replacement using a fixed seed of 0 and consumed in a single fixed order shared across all variants, so that a projector-only run and its joint counterpart see the exact same stream of data.

\paragraph{Sequence construction.} Prompts are formatted with one of the system messages shown in Table~\ref{tab:systempool}, followed by point tokens inserted between \texttt{<|pc3d\_start|>} and \texttt{<|pc3d\_end|>} tags, and ending with a specific question from our training dataset. The labels on the prompt tokens are masked with $-$100 so only the model response tokens contribute to our loss. Sequences are truncated at 512 tokens, although this limit is never reached in practice.

\begin{table}[h]
\centering
\caption{The set of system message paraphrases, drawn uniformly at random during training.}
\label{tab:systempool}
\begin{tabular}{l}
\toprule
System message \\
\midrule
Answer the question or follow the instruction regarding the given 3D object. \\
Answer the following question or carry out the instruction about the provided 3D object. \\
Given a 3D object, respond to the question or instruction about it. \\
Consider the 3D object and answer the question or follow the instruction that follows. \\
Respond to the question or instruction concerning the presented 3D object. \\
Using the given 3D object, answer the question or complete the instruction. \\
You are given a 3D object; answer the question or follow the instruction about it. \\
Examine the 3D object and answer the accompanying question or instruction. \\
Provide an answer to the question or complete the instruction about the given 3D object. \\
Based on the 3D object shown, answer the question or follow the instruction. \\
Address the question or instruction about the provided 3D object. \\
\bottomrule
\end{tabular}
\end{table}

\subsection{Optimization}

All parameters are optimized with AdamW under the token-averaged loss objective $\mathcal{L}(\theta) = - \frac{1}{T} \sum{\log{P_{\theta}(x_{t+1} | x_{\leq t})}}$. Projector-only training optimizes a single parameter group while joint training optimizes two groups under one optimizer instance, with separate projector and LoRA learning rates. The schedule is constant with no warmup and zero weight decay in both regimes. As the paper primarily concerns comparisons between projector-only and joint training regimes, we deliberately choose a plain config to avoid excessive hyperparameter tuning for the sake of maximizing performance. The projector learning rate is the one setting that varies across configurations, chosen empirically using a small sweep. Table \ref{tab:hparams} shows all hyperparameter configurations; we generally followed the learning rates in PointLLM-V2 \citep{pointllmv2}, though we modified the Llama8B learning rates as the original values led to underperforming MLLMs for both projector-only and joint training.

\begin{table}[h]
\centering
\caption{Training and evaluation hyperparameters, shared across every configuration except the learning rates.}
\label{tab:hparams}
\setlength{\tabcolsep}{6pt}
\renewcommand{\arraystretch}{1.12}
\begin{tabular}{@{}ll@{}}
\toprule
\textbf{Hyperparameter} & \textbf{Value} \\
\midrule
Optimizer                            & AdamW, $\beta = (0.9, 0.999)$, $\epsilon = 10^{-8}$ \\
Learning-rate schedule               & constant, no warmup \\
Projector learning rate              & $2 \times 10^{-3}$ ($1 \times 10^{-3}$ for Llama8B) \\
Weight decay                         & $0.0$ \\
Gradient clipping                    & global norm $1.0$ \\
Micro-batch / accumulation / effective & $12$ / $2$ / $24$ \\
Max response length                  & $512$ tokens \\
Mixed precision                      & \texttt{bfloat16} autocast \\
Random seed                          & $0$ \\
Hardware                             & $1 \times$ NVIDIA A100-80GB \\
Gradient checkpointing               & enabled \\
\midrule
LoRA rank / $\alpha$ / dropout       & $16$ / $32$ / $0.05$ \\
LoRA target modules                  & \texttt{all-linear} \\
LoRA learning rate      & $2 \times 10^{-5}$ ($2 \times 10^{-4}$ for Llama8B) \\
\bottomrule
\end{tabular}
\end{table}

\subsection{Throughput and Samples Seen}

Table~\ref{tab:samples-over-time} shows the total number of training samples seen by each MLLM variant after 16 GPU hours. Given the same language model backbone, projector-only training processes samples at approximately twice the rate of joint training. This ratio holds near $2 \times$ across all three pairings, and is due to joint training having to track and backpropagate into LoRA adapter weights.

A secondary ordering is visible within each training regime, where each backbone has a different relative training rate. Qwen3.5-4B is the fastest, followed by Llama-3.1-8B-Instruct and Qwen3.5-9B, as expected when the frozen forward pass dominates cost.

\begin{table}[h]
    \centering
    \caption{Total number of training samples seen by each MLLM variant, average throughput, and average step rate after 16 GPU hours.}
    \label{tab:samples-over-time}
    \begin{tabular}{lccc}
        \toprule
        ID & Total samples & Avg. throughput (h$^{-1}$) & Avg. step rate (h$^{-1}$) \\
        \midrule
        P-Llama8B & 432,936 & 27,056 & 1,127  \\
        J-Llama8B & 200,640 & 12,537 &   522  \\
        P-Qwen4B  & 480,768 & 30,045 & 1,252  \\
        J-Qwen4B  & 245,592 & 15,346 &   639  \\
        P-Qwen9B  & 325,176 & 20,321 &   847  \\
        J-Qwen9B  & 174,720 & 10,918 &   455  \\
        \bottomrule
    \end{tabular}
\end{table}

\newpage

\section{Evaluation Procedure}
\label{app:eval-procedure}

\subsection{Generative Zero-Shot Classification}

\paragraph{ModelNet40.} We use PointLLM's released test file, which contains the standard 2,468 object test split, already farthest-point sampled to 8192 points. We keep the XYZ channels and discard the surface normals. ModelNet40 is geometry-only CAD data with no color, so every point is assigned $\mathrm{RGB}=-1$ (black) under our color convention.

\paragraph{Objaverse.} We use PointLLM's 3,000 object held-out benchmark with their brief-description ground truth as the reference answer. There is no fixed label set, since the reference is a short free-form caption for each object. The point clouds are the release's own 8192-point colored point clouds, extracted per object from the released data.

\paragraph{OmniObject3D.} The ground truth is PointLLM-R's released brief-description validation set, which contains 5,900 objects over 216 categories. The official OmniObject3D point clouds carry no color, so we transfer color onto them from the raw textured scans, keeping the point positions byte-identical and changing only the color channels.

\paragraph{Common processing.} At evaluation time, each point cloud is subsampled to 8192 points, its coordinates are normalized to the unit sphere, and color is passed in $[-1, 1]$. Generations are greedy with a 256-token budget, and each benchmark is run under both variants of PointLLM's classification prompts: \texttt{"This is an object of"} (C) and \texttt{"What is this?"} (I).

\paragraph{Judge protocol.} Generations are scored by GPT-5.6 Luna~\citep{gpt56} (\texttt{gpt-5.6-luna}), with a 96-token visible reply budget. Two different rubrics are used, shown in Figures \ref{fig:judgeprompts-cls-obj} and \ref{fig:judgeprompts-cls-mn40}, depending on whether a fixed set of classes is available. On Objaverse and OmniObject3D, the judge sees the reference caption and simply decides whether there is a match. On ModelNet40, however, the judge maps the free-form generation onto one of the 40 classes and returns the closest one, without seeing the ground truth answer.

\subsection{Object Captioning}

\paragraph{Evaluation set.} We use the same 3,000 Objaverse objects as the classification benchmark, each paired with the 8 rendered views shipped by Cap3D \citep{cap3d}.

\paragraph{Model prompt.} We use PointLLM's captioning prompt verbatim: ``Caption this 3D model in detail.'' Captions are decoded greedily with a 256-token budget, identical to the classification evaluations.

\paragraph{Scoring rubric.} Each caption is scored against the 8 views by an LLM judge with no reference caption; the images are the only ground truth. The judge identifies the key visual aspects of the object (category, color, shape, usage, material) and returns a correctness score $C$ and a hallucination score $H$ as \emph{point counts}, not percentages: one correctness point per correctly stated aspect (fractional credit in $[0,1]$ for partial matches), and one hallucination point per aspect the caption asserts that the images contradict. Precision is computed similarly to PointLLM-V2, as a ratio of the total correctness score to the sum of the correctness and hallucination scores over the entire evaluation set.

\paragraph{Judges.} We use three frontier LLMs as judges: GPT-5.6 Luna \citep{gpt56} (\texttt{gpt-5.6-luna}), Claude Haiku 4.5 \citep{haiku45} (\texttt{claude-haiku-4-5-20251001}), and Gemini 3.5 Flash-Lite \citep{gemini35} (\texttt{gemini-3.5-flash-lite}), each with a 200-token visible reply budget. To account for variance between these judges, we average their precision scores together, as is done in PointLLM-V2. All three judges receive an identical single user message (no system message), shown in Figure~\ref{fig:judgeprompts-cap}.

\newpage

\begin{figure}[H]
\caption{The judge prompt for generative zero-shot classification on Objaverse and OmniObject3D. Placeholders in braces are substituted at evaluation time; \{gold\} and \{generation\} refer to the ground-truth caption and the model's response, respectively.}
\label{fig:judgeprompts-cls-obj}
\small
\centering
\begin{tcolorbox}[
  enhanced,
  colback=white,
  colframe=gray!70!black,
  boxrule=1.2pt,
  arc=4mm,
  left=8mm,
  right=8mm,
  top=8mm,
  bottom=6mm,
  before skip=6pt,
  after skip=10pt,
]
Analyze two sentences and determine if they're referring to the same general object or concept, focusing on the type of object, not attributes such as color, size, or shape. Respond with 'T' if they refer to the same thing and 'F' if not. Also, provide a brief rationale (no more than 20 words) for your judgment. \\

Example: \\
Input: 1. A black and brown colored gun. 2. The 3D object is a representation of a futuristic, high-tech gun crafted from a glossy black material. Distinctive features include its metallic handrail, giving an impression of a robust mechanized design. The gun, possibly used in a sci-fi or futuristic setting, denotes advanced technology and might include functionalities such as voice recognition, aiming systems, or biometric triggers. \\
Output: T\#Both refer to a gun. \\
Input: 1. A yellow and white fish with black stripes and fins. 2. This is a 3D model of a vibrant, polka-dotted toy fish that is predominantly orange on the body, shifting to white on the belly. The toy has dark brown spots that enhance its appearance, potentially mimicking the natural patterns found on real-life fish. It's an ideal object for educational purposes, helping to introduce children to marine life, as well as serving as a playful item in a playroom or nursery. \\
Output: T\#Both refer to a fish. \\
Input: 1. A white cartoon scorpion with eight legs. 2. This is a 3D object model representing a cartoon version of a rare type of spider. The entire model is rendered in white, which highlights its unique and exaggerated characteristics such as multiple legs and a funnel-like body. Its cartoonish appeal makes it more appealing to a younger audience, and it could possibly be used in animations or educational materials to teach children about spiders in a less intimidating way. \\
Output: F\#One is a scorpion and the other is a spider. \\

Now, analyze the following: \\
Input: 1. \{gold\} 2. \{generation\} \\
Output:
\end{tcolorbox}
\end{figure}

\begin{figure}[H]
\caption{The judge prompt for generative zero-shot classification on ModelNet40. The placeholder \{generation\} refers to the model's response.}
\label{fig:judgeprompts-cls-mn40}
\small
\centering
\begin{tcolorbox}[
  enhanced,
  colback=white,
  colframe=gray!70!black,
  boxrule=1.2pt,
  arc=4mm,
  left=8mm,
  right=8mm,
  top=8mm,
  bottom=6mm,
  before skip=6pt,
  after skip=10pt,
]
Given the following free-form description of a 3D object, please determine the most probable class index from the following 40 available categories, even if the description doesn't clearly refer to any one of them. Make your best-educated guess based on the information provided. If the description already contains a valid index, then the index should be selected. If it contains more than one valid index, then randomly select one index (specify your reason). If there is no valid index and it cannot be inferred from the information, return ``-1\#NA\#Cannot infer''. \\

Categories: \\
0: airplane
1: bathtub
2: bed
3: bench
4: bookshelf
5: bottle
6: bowl
7: car
8: chair
9: cone
10: cup
11: curtain
12: desk
13: door
14: dresser
15: flower\_pot
16: glass\_box
17: guitar
18: keyboard
19: lamp
20: laptop
21: mantel
22: monitor
23: night\_stand
24: person
25: piano
26: plant
27: radio
28: range\_hood
29: sink
30: sofa
31: stairs
32: stool
33: table
34: tent
35: toilet
36: tv\_stand
37: vase
38: wardrobe
39: xbox \\
Reply with the format of ``index\#class\#short reason (no more than 10 words)''. \\

Examples: \\
Input: This is a 3D object model of a cartoon white truck. \\
Output: 7\#car\#Closest match to ``car'' in categories. \\
Input: A green leaf in a flower pot. \\
Output: 26\#plant\#The primary subject ``leaf'' directly indicates a plant. \\
Input: It's difficult to determine the exact type of this object due to insufficient details. But it seems to be like a piece of furniture. \\
Output: 33\#table\#Randomly select one kind of furniture from the list. \\
Input: I cannot determine the specific type of the object without additional information or context. \\
Output: -1\#NA\#Cannot infer. \\

Now analyze the following: \\
Input: \{generation\} \\
Output:
\end{tcolorbox}
\end{figure}

\begin{figure}[H]
\caption{The judge prompt for object captioning on Objaverse. Placeholders in braces are substituted at evaluation time; \{generation\} refers to the model's generated caption. The judge additionally receives 8 rendered views of the object alongside this text.}
\label{fig:judgeprompts-cap}
\small
\centering
\begin{tcolorbox}[
  enhanced,
  colback=white,
  colframe=gray!70!black,
  boxrule=1.2pt,
  arc=4mm,
  left=8mm,
  right=8mm,
  top=8mm,
  bottom=6mm,
  before skip=6pt,
  after skip=10pt,
]
You are provided with 8 images of an object taken from different angles. Your task is to evaluate a model-generated caption based on the visual information in these images. Identify the key aspects (e.g., category, color, shape, usage, material) from the images and calculate the percentage of these aspects that are correctly mentioned or partially matched in the model-generated caption. Assign correctness points for each distinct correct attribute. Partial correctness should be graded on a scale of 0 to 1 depending on accuracy, with similar concepts considered for partial scores. Each aspect contributes equally to the final score, with no extra penalties for repeated inaccuracies of the same attribute. Also, assign hallucination points for incorrect details in the model output, with one point per incorrect attribute. Repetitive inaccuracies based on one attribute should incur only a single hallucination point. However, if a detail is described uncertainly or speculatively (e.g., using phrases like 'probably for', 'possibly', 'resemble' or 'indicating'), assign fewer or no hallucination points. Provide your score and a short justification (less than 25 words) in the format of ``Output: score\#your reason'' \\

Examples (corresponding images omitted): \\
Example 1 \\
Model: The object is a geometric shape with a complex design featuring interlocking blue and white elements. It appears to be a three-dimensional structure with angular and rectangular cutouts, creating a visually intricate form. The blue elements form the primary color scheme, contrasted by white areas that highlight the depth and complexity of the design. \\
Correctness: 4.5\#Correct identification of geometric shape, colors, angular and rectangular cutouts, and three-dimensional structure. \\
Hallucination: 0\#No incorrect details. \\
Example 2 \\
Model: The object is a green, cartoonish character with a rounded body and two short limbs protruding from the sides. It has two antennae on top of its head and a pair of eyes on the front. The overall shape is smooth and symmetrical, resembling a playful or animated figure. \\
Correctness: 3.5\#The object is green, cartoonish, and has a rounded body with two limbs. \\
Hallucination: 2.0\#No antennae or eyes are visible, shape is not entirely smooth. \\
Model: \{generation\} \\

Now give your score and reason without adding extra content. Reply with exactly two lines, and express both scores as point counts (a number of attributes, decimals allowed for partial credit) --- not percentages: \\
Correctness: <points>\#<reason> \\
Hallucination: <points>\#<reason>
\end{tcolorbox}
\end{figure}

\newpage

\section{Additional Results}
\label{app:add-results}

Tables \ref{tab:results-3dund-extension} and \ref{tab:caption-16h-extension} are extensions of Tables \ref{tab:results-3dund} and \ref{tab:caption-16h} respectively, with additional rows for more existing models, joint variants at 16 hours, the curriculum learning ablation, and joint variants without LoRA adapters loaded. Table \ref{tab:acc-over-time-numbers} reports the exact values plotted in Figure \ref{fig:acc-over-time} comparing projector-only and joint training runs.

\begin{table}[H]
    \caption{Generative 3D object classification results on ModelNet40 (M40.), Objaverse (Obj.), and OmniObject3D (Omni.) under a zero-shot setting. Extension of Table \ref{tab:results-3dund}.}
    \label{tab:results-3dund-extension}
    \centering
    \begin{tabular}{lcccccc}
    \toprule
    Model & M40. (I) & M40. (C) & Obj. (I) & Obj. (C) & Omni. (I) & Omni. (C) \\
    \midrule
    ShapeLLM-7B \cite{shapellm}  & 18.76 & 17.95 & 30.23 & 31.37 & 15.28 & 18.75 \\
    ShapeLLM-13B \cite{shapellm} & 22.45 & 21.60 & 40.67 & 38.90 & 28.66 & 33.33 \\
    PointLLM-7B \cite{pointllm}  & 54.94 & 55.55 & 59.83 & 60.43 & 34.78 & 35.70 \\
    PointLLM-13B \cite{pointllm} & 57.66 & 56.81 & 61.73 & 60.90 & 34.63 & 34.01 \\
    PointLLM-R \cite{pointllm-r} & 62.28 & 62.64 & 65.23 & 65.40 & 38.81 & 38.69 \\
    MiniGPT-3D \cite{minigpt3d}  & 63.41 & 62.84 & 63.97 & 63.00 & 41.10 & 39.27 \\
    \midrule
    P-Llama8B & 61.95 & 63.57 & 65.60 & 62.33 & 49.19 & 40.69 \\
    J-Llama8B & 49.72 & 57.37 & 66.80 & 59.53 & 42.34 & 37.17 \\
    J-NoLoRA-Llama8B & 36.02 & 40.24 & 58.90 & 51.07 & 39.62 & 40.69 \\
    \addlinespace[2pt]
    P-Qwen4B & 59.56 & 59.56 & 60.90 & 60.83 & 41.07 & 40.25 \\
    J-Qwen4B & 58.27 & 59.60 & 71.00 & 68.80 & 48.17 & 47.32 \\
    J-NoLoRA-Qwen4B  & 54.94 & 53.24 & 58.73 & 57.97 & 27.08 & 27.49 \\
    \addlinespace[2pt]
    P-Qwen9B & 59.28 & 62.80 & 74.07 & 67.77 & 48.33 & 38.76 \\
    J-Qwen9B & 52.59 & 58.75 & 75.23 & 66.23 & 51.32 & 44.86 \\
    J-NoLoRA-Qwen9B  & 51.62 & 52.63 & 49.13 & 56.93 & 24.51 & 28.74 \\
    \addlinespace[2pt]
    Curriculum-Qwen4B & 53.48 & 54.34 & 74.80 & 75.63 & 46.08 & 46.73 \\
    \bottomrule
    \end{tabular}
\end{table}

\begin{table}[H]
    \caption{3D object captioning results on Objaverse. Extension of Table \ref{tab:caption-16h}.}
    \label{tab:caption-16h-extension}
    \centering
    \begin{tabular}{lccccccccc}
      \toprule
       & \multicolumn{3}{c}{GPT-5.6 Luna}
       & \multicolumn{3}{c}{Claude Haiku 4.5}
       & \multicolumn{3}{c}{Gemini 3.5 Flash-Lite} \\
      \cmidrule(lr){2-4} \cmidrule(lr){5-7} \cmidrule(lr){8-10}
      Model & C & H $\downarrow$ & P
            & C & H $\downarrow$ & P
            & C & H $\downarrow$ & P \\
      \midrule
      ShapeLLM-7B \citep{shapellm}  & 2.23 & 1.52 & 59.47 & 1.65 & 1.46 & 53.15 & 1.03 & 1.69 & 37.85 \\
      ShapeLLM-13B \citep{shapellm} & 2.68 & 1.84 & 59.32 & 2.04 & 1.68 & 54.93 & 1.39 & 2.02 & 40.81 \\
      PointLLM-7B \citep{pointllm}  & 4.26 & 2.12 & 66.75 & 3.37 & 1.95 & 63.30 & 2.75 & 1.99 & 58.08 \\
      PointLLM-13B \citep{pointllm} & 4.35 & 2.10 & 67.43 & 3.47 & 1.91 & 64.46 & 2.82 & 1.92 & 59.49 \\
      PointLLM-R \citep{pointllm-r} & 3.38 & 1.18 & 74.10 & 2.50 & 1.10 & 69.42 & 2.48 & 1.17 & 67.94 \\
      MiniGPT-3D \citep{minigpt3d}  & 4.51 & 2.01 & 69.16 & 3.19 & 1.93 & 62.32 & 2.70 & 1.95 & 58.07 \\
      \midrule
      P-Llama8B    & 5.61 & 2.17 & 72.12 & 3.76 & 1.75 & 68.19 & 3.29 & 1.87 & 63.75 \\
      J-Llama8B   & 6.09 & 2.35 & 72.15 & 4.01 & 1.95 & 67.28 & 3.41 & 2.11 & 61.79 \\
      J-NoLoRA-Llama8B & 3.14 & 2.96 & 51.48 & 2.35 & 2.08 & 52.96 & 1.69 & 2.31 & 42.15 \\
      \addlinespace[2pt]
      P-Qwen4B     & 6.41 & 1.83 & 77.81 & 4.28 & 1.57 & 73.11 & 3.72 & 1.65 & 69.22 \\
      J-Qwen4B    & 6.73 & 1.97 & 77.33 & 4.43 & 1.67 & 72.67 & 3.91 & 1.71 & 69.52 \\
      J-NoLoRA-Qwen4B  & 5.44 & 4.23 & 56.26 & 4.30 & 2.53 & 62.97 & 2.92 & 3.19 & 47.80 \\
      \addlinespace[2pt]
      P-Qwen9B     & 6.58 & 1.95 & 77.19 & 4.30 & 1.68 & 71.90 & 3.76 & 1.71 & 68.72 \\
      J-Qwen9B    & 6.83 & 2.02 & 77.19 & 4.46 & 1.73 & 72.00 & 3.92 & 1.80 & 68.52 \\
      J-NoLoRA-Qwen9B  & 5.14 & 4.74 & 52.02 & 3.84 & 3.02 & 56.00 & 2.58 & 3.65 & 41.46 \\
      \addlinespace[2pt]
      Curriculum-Qwen4B & 6.52 & 1.95 & 76.96 & 4.36 & 1.67 & 72.33 & 3.76 & 1.74 & 68.38 \\
      \bottomrule
    \end{tabular}
\end{table}

\begin{table}[H]
\centering
\caption{Generative zero-shot classification accuracy against wall-clock hours. This is calculated as the average of the I and C variants of the ModelNet40 and Objaverse benchmarks, directly comparable to Figure~\ref{fig:acc-over-time}.}
\label{tab:acc-over-time-numbers}
\setlength{\tabcolsep}{6pt}
\begin{tabular}{lccccc@{\hskip 12pt}ccccc}
\toprule
& \multicolumn{5}{c}{M40. (Avg.)} & \multicolumn{5}{c}{Obj. (Avg.)} \\
\cmidrule(lr){2-6} \cmidrule(l){7-11}
Model & 2\,h & 4\,h & 8\,h & 12\,h & 16\,h & 2\,h & 4\,h & 8\,h & 12\,h & 16\,h \\
\midrule
P-Qwen4B & 54.23 & 54.76 & 58.33 & 63.53 & 59.56 & 58.10 & 62.52 & 67.25 & 64.90 & 60.87 \\
J-Qwen4B & 46.84 & 51.64 & 51.03 & 52.09 & 58.93 & 58.20 & 66.50 & 73.28 & 75.05 & 69.90 \\
\addlinespace[2pt]
P-Qwen9B & 51.38 & 53.99 & 57.03 & 61.06 & 61.04 & 57.47 & 63.07 & 67.80 & 67.35 & 70.92 \\
J-Qwen9B & 39.49 & 46.84 & 54.52 & 58.53 & 55.67 & 51.20 & 60.48 & 71.08 & 67.90 & 70.73 \\
\addlinespace[2pt]
P-Llama8B & 53.85 & 57.22 & 57.80 & 56.14 & 62.76 & 51.48 & 57.58 & 66.10 & 70.07 & 63.97 \\
J-Llama8B & 45.50 & 48.28 & 47.49 & 56.62 & 53.55 & 47.80 & 54.02 & 62.28 & 65.23 & 63.17 \\
\addlinespace[2pt]
\bottomrule
\end{tabular}
\end{table}

\newpage

\section{Example Generations}
\label{app:qual-evals}

Table \ref{tab:qual-captions-appendix} shows captions generated by P-Llama8B and J-Llama8B at 16 GPU hours. They are often able to correctly identify shape, color, and object semantics, such as what the object is, though in some challenging settings (e.g., low-poly objects) they fail to identify the object class correctly while still identifying textural details accurately.

\begin{table}[H]
    \caption{Example generations on long captioning tasks. The MLLMs are given the prompt ``Caption this 3D model in detail.'' All prompts are independently provided to the MLLMs. The answers shown in this table were generated by P-Llama8B and J-Llama8B at 16 hours of training, as well as the two PointLLM models.}
    \label{tab:qual-captions-appendix}
    \centering
    \small
    \begin{tabular}{lp{0.4\linewidth}p{0.4\linewidth}}
    \toprule
    Object & \includegraphics[width=2.6cm]{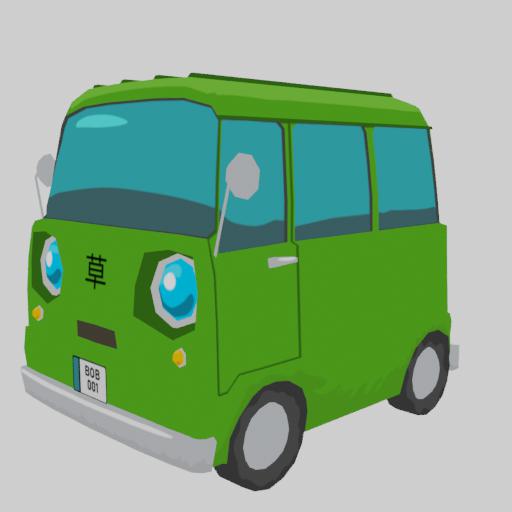} \includegraphics[width=2.6cm]{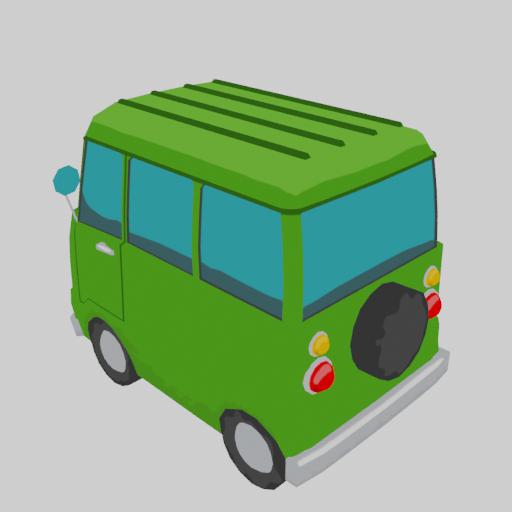} & \includegraphics[width=2.6cm]{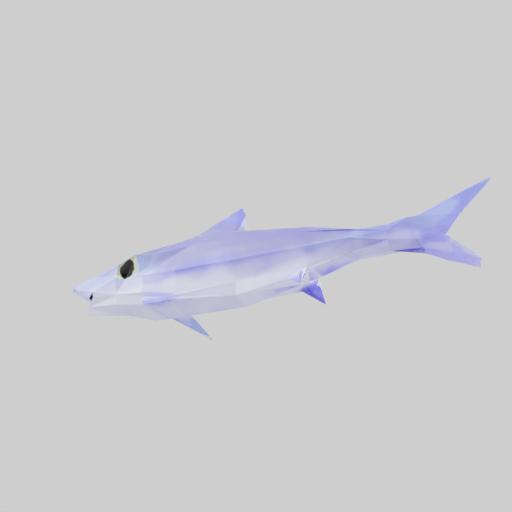} \includegraphics[width=2.6cm]{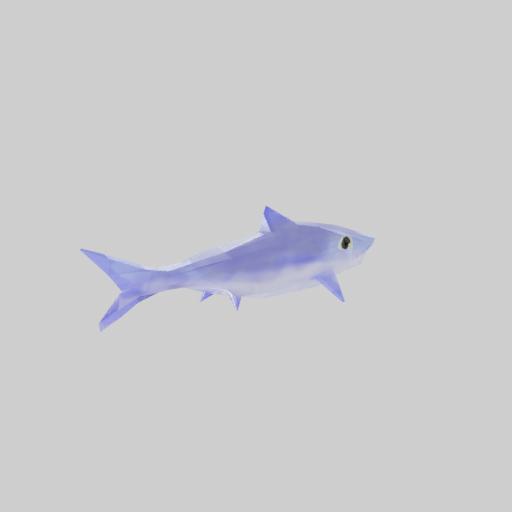} \\
    \midrule
    PointLLM-7B & The 3D model represents a distinctive, green toy car that stands out due to its unconventional feature - a cornered driving unit. Designed for play, this model is aesthetically appealing with an eye-catching bright green color that would attract young children. The unique driving unit located at the corner provides an interesting twist to the traditional car design, promoting creative play and imaginative scenarios. However, despite its playful design, it is still a toy and should be handled with care. & The 3D object is a lively representation of a blue whale, which is the largest mammal on Earth. It is depicted in a playful cartoon style with an eye-catching cobalt-blue hue. The whale features proportional body parts, including a small dorsal fin, enormous mouth, and flat, wide head. It seems to be designed for entertainment or educational purposes, possibly for children, given its cartoon-like attributes and the significance of the whale in marine life. \\
    PointLLM-13B & The 3D object model represents a small toy car, with the primary colour being a vibrant green. It features four black wheels that contrast well with the green body. Two red lights are situated at the front, possibly to mimic the headlights of an actual car. This model, being a toy, is typically used for children's play, aiding in their motor skill development and imagination. The car's simplistic design might be suitable for younger age groups. & This is a 3D model of a blue whale, the largest mammal on Earth, depicted in a cartoon style. The whale is distinguished by its small dorsal fin, enormous mouth, and flat, wide head. Despite the cartoon aesthetic, the model accurately represents the unique features of a blue whale. It could be used for educational purposes or in animated movies or documentaries. \\
    \textbf{P-Llama8B} & The object is a green and white van with a boxy shape, featuring a prominent front grille, round headlights, and a rear window. It has a flat roof and a side door with a handle. The van has a distinctive green and white color scheme with black accents. The wheels are circular and black, and the vehicle has a distinctive front bumper. & The object is a sleek, streamlined shark with a pointed snout and a series of sharp, triangular fins along its back. The body is predominantly blue, with a lighter underside and a distinctive pattern of darker spots. The fins are a lighter shade, and the tail is long and narrow, tapering to a sharp point. The shark has a prominent dorsal fin and a smaller pectoral fin, with a series of gill slits visible along the sides. \\
    \textbf{J-Llama8B} & The object is a green vehicle with a boxy design, featuring a large windshield, a rear window, and a front grille. It has a roof rack and a spare tire mounted on the rear. The vehicle appears to be a van, with a prominent front bumper and a rear section that includes a license plate. The wheels are circular, and the vehicle has a simple, utilitarian appearance. & The object is a stylized representation of a fish, characterized by a streamlined body with a pointed snout and a forked tail. The body is primarily blue with a white underbelly, and the fins are white. The fish has a smooth, abstract appearance with a minimalistic design. \\
    \bottomrule
    \end{tabular}
\end{table}

\end{document}